\documentclass[conference]{IEEEtran}
\IEEEoverridecommandlockouts
\usepackage{cite}
\usepackage{amsmath,amssymb,amsfonts}
\usepackage{algorithmic}
\usepackage{graphicx}
\usepackage{textcomp}
\usepackage[titlenumbered,ruled]{algorithm2e}
\usepackage{graphicx}
\usepackage{subcaption}
\usepackage{graphicx}  
\usepackage{subcaption}  
\usepackage{comment}
\usepackage{algorithmic}
\usepackage{graphicx}
\usepackage{textcomp}
\usepackage{hyperref}  
\usepackage{amsmath}
\usepackage{tabularx}
\usepackage{booktabs}  

\usepackage{xcolor}
\usepackage{caption}
\usepackage{subcaption}
\usepackage{tabularx} \usepackage{xcolor}
\def\BibTeX{{\rm B\kern-.05em{\sc i\kern-.025em b}\kern-.08em
    T\kern-.1667em\lower.7ex\hbox{E}\kern-.125emX}}
\begin{document}

\title{
UAN: Unsupervised Adaptive Normalization
}

\author{\IEEEauthorblockN{1\textsuperscript{st} Bilal Faye}
\IEEEauthorblockA{\textit{LIPN, UMR CNRS 7030} \\
Villetaneuse, France \\
faye@lipn.univ-paris13.fr}
\and
\IEEEauthorblockN{2\textsuperscript{nd} Hanane Azzag}
\IEEEauthorblockA{\textit{LIPN, UMR CNRS 7030} \\
Villetaneuse, France\\
azzag@univ-paris13.fr}
\and
\IEEEauthorblockN{3\textsuperscript{rd} Mustapha Lebbah}
\IEEEauthorblockA{\textit{DAVID laboratory} \\
Versailles, France\\
mustapha.lebbah@uvsq.fr}
\and
\IEEEauthorblockN{4\textsuperscript{th} Fangchen FANG}
\IEEEauthorblockA{\textit{L2TI laboratory} \\
Villetaneuse, France \\
fangchen.feng@univ-paris13.fr}
}
\maketitle

\begin{abstract}
Deep neural networks have become a staple in solving intricate problems, proving their mettle in a wide array of applications. However, their training process is often hampered by shifting activation distributions during backpropagation, resulting in unstable gradients. Batch Normalization (BN) addresses this issue by normalizing activations, which allows for the use of higher learning rates. Despite its benefits, BN is not without drawbacks, including its dependence on mini-batch size and the presumption of a uniform distribution of samples. To overcome this, several alternatives have been proposed, such as Layer Normalization, Group Normalization, and Mixture Normalization. These methods may still struggle to adapt to the dynamic distributions of neuron activations during the learning process. To bridge this gap, we introduce Unsupervised Adaptive Normalization (UAN), an innovative algorithm that seamlessly integrates clustering for normalization with deep neural network learning in a singular process.
UAN executes clustering using the Gaussian mixture model, determining parameters for each identified cluster, by normalizing neuron activations. These parameters are concurrently updated as weights in the deep neural network, aligning with the specific requirements of the target task during backpropagation. This unified approach of clustering and normalization, underpinned by neuron activation normalization, fosters an adaptive data representation that is specifically tailored to the target task. This adaptive feature of UAN enhances gradient stability, resulting in faster learning and augmented neural network performance.
UAN outperforms the classical methods by adapting to the target task and is effective in classification, and domain adaptation. 
We release our code at \href{https://github.com/b-faye/unsupervised-context-normalization}{github repository}
\end{abstract}


\section{Introduction}
Deep neural networks have shown significant advancements in various domains. However, they still face specific challenges, particularly in terms of accelerating learning and improving performance. The diverse nature of data often requires pre-training methods to normalize them to the same scale and eliminate biases. In a neural network with a single layer, normalizing input data accelerates convergence, thus enhancing performance~\cite{lecun2002efficient}. This approach, however, is limited as contemporary neural networks consist of multiple layers. Input data is connected only to the first layer of the neural network, which does not guarantee that subsequent layers will benefit from input data normalization due to weight updates during the backward pass. To address this, normalization approaches have been proposed for layers not directly connected to input data, including activation normalization, weight normalization, and gradient normalization. Batch Normalization (BN) introduced by Ioffe and Szegedy~\cite{ioffe2015batch} normalizes the output of each neural network layer to have a zero mean and unit variance property. This approach stabilizes gradient variation, accelerating convergence and improving deep neural network performance. Despite its effectiveness, BN has limitations such as mini-batch dependency, restricting its utility for small batches, and the weak assumption that samples within the mini-batch are from the same distribution, which is not always true, especially with real-world data.

Layer Normalization (LN)~\cite{ba2016layer}, Instance Normalization (IN)~\cite{ulyanov2016instance}, and Group Normalization (GN)~\cite{wu2018group} were proposed to address mini-batch dependency issues. Mixture Normalization (MN)~\cite{kalayeh2019training} presents an approach based on the assumption of Gaussian mixture models (GMM) on input data to overcome the distribution assumption made by BN. MN employs a dual-stage process for normalizing neuron activations: (i)Expectation-Maximization (EM) algorithm~\cite{Dempster77maximumlikelihood} is initially used to deduce parameters for each mixture component; (ii) in the training of deep neural networks, samples linked to the same component are normalized utilizing the parameters specific to their component. This approach has demonstrated quicker convergence and enhanced performance in neural networks, particularly in tasks of supervised classification. However, MN's dependence on the EM algorithm as a preprocessing step could be a limiting factor due to its computational intensity. Additionally, MN's static parameters might not be sufficiently adaptive to the shifting distributions of neuron activations during the learning phase.

To tackle this challenge, we introduce Unsupervised Adaptive Normalization (UAN), a streamlined normalization approach. UAN is grounded in the assumption that data are governed by a Gaussian mixture model in a latent space. The objective is to discover this space during the deep neural network learning process by normalizing neuron activations. The parameters of each mixture component called "cluster", are estimated as weights of the deep neural network during the backpropagation phase, enabling the construction of mixture components tailored to the target task addressed by the deep neural network. This one-pass approach provides an adaptive data representation based on the clustering methodology, enhancing separability and facilitating deep neural network learning. 
Notably, UAN demonstrates superior results in terms of convergence and deep neural network performance when compared to BN and MN.

In summary, the main contributions of this work are as follows:
\begin{itemize}
   \item \textbf{Simultaneous clustering for normalization and deep Learning:}
   Unsupervised Adaptive Normalization(UAN) integrates clustering and deep neural network learning in a single step. By normalizing neuron activations based on the Gaussian mixture model assumption, UAN estimates parameters for each mixture component, improving data representation for the target task and boosting gradient stability.

   \item \textbf{Efficiency and complexity reduction:}
   UAN stands out by simplifying the architecture and eliminating the preprocessing clustering phase using EM used in Mixture Normalization. UAN adapts normalization to the target task, determining mixture component parameters during backpropagation as model weights. This streamlined approach demonstrates superior convergence and performance compared to BN and MN.

   \item \textbf{Enhanced convergence and performance:}
   UAN surpasses Batch Normalization and Mixture Normalization regarding convergence and deep neural network performance. Its adaptive data representation, achieved through simultaneous clustering and learning, enhances separability and accelerates the learning process. 
\end{itemize}

\section{State of the art}
\label{section:state_art}
To ensure clear understanding and consistency in our model, and to facilitate comparison with existing work, we adopt the same notations as those used in  \cite{kalayeh2019training}. Let $x \in \mathbb R^{N\times C\times H\times W}$ neurons activations in a convolutional neural network layer, where $N$, $C$, $H$, and $W$ represent the batch, channel, height, and width axes, respectively. Batch Normalization (BN)~\cite{ioffe2015batch} standardizes a mini-batch $B = \{x_{1:m}: m \in [1, N]\times [1, H] \times [1, W]\}$ of $m$ samples with $x$ flattened across all axes except the channel. The standardization is expressed as:
\begin{equation}
\label{bn_equation}
\hat{x}_{i} = \frac{x_{i}-\mu_B}{\sqrt{\sigma^2_B+\epsilon}},
\end{equation}
where $\mu_B = \frac{1}{m} \sum_{i=1}^m x_{i}$, $\sigma^2_B = \frac{1}{m}\sum_{i=1}^m (x_{i}-\mu_B)^2$ are respectively the mean and variance. $\epsilon > 0$ is a small number to prevent numerical instability. To mitigate the constraints imposed by standardization, learnable parameters $\gamma$ and $\beta$ are introduced, transforming activations as follows:
\begin{equation}
\Tilde{x}_{i} = \gamma\hat{x}_{i} + \beta.
\end{equation}
During inference, population statistics are essential for deterministic inference. They are computed by averaging over training iterations:
\begin{equation}
\begin{cases}
\hat{\mu} = (1-\lambda)\hat{\mu} + \lambda \mu_B \\
\hat{\sigma}^2 = (1-\lambda)\hat{\sigma}^2 + \lambda \sigma^2_B.
\end{cases}
\end{equation}
If the mini-batch samples come from the same distribution, the transformation in equation~\eqref{bn_equation} yields a distribution with zero mean and unit variance. This constraint stabilizes activation distribution, aiding training.\newline
\indent The mini-batch-wise approach allows for a more suitable data representation, enabling faster processing and enhancing deep neural network performance in terms of convergence. Despite its efficacy, BN's performance is contingent on mini-batch size, and the disparity between training and inference hinders its application in intricate networks (e.g., recurrent neural networks). To address these challenges and handle parameter estimation on unrelated samples, numerous variants and alternative methods have been proposed.\newline
\indent Various extensions to batch normalization (BN) have emerged, including Layer Normalization (LN)~\cite{ba2016layer}, Instance Normalization (IN)~\cite{ulyanov2016instance}, Group Normalization (GN)~\cite{wu2018group}, and Mixture Normalization (MN)~\cite{kalayeh2019training}. These alternatives share a common transformation $x \rightarrow \hat{x}$, applied to the flattened $x$ across the spatial dimension $L = H \times W$, expressed as:
\begin{equation}
v_{i} = x_{i} - \mathbb{E}_{B_i}(x), \
\hat{x}_{i} = \frac{v_{i}}{\sqrt{\mathbb{E}_{B_i}(v^2)+\epsilon}},
\label{general_transform}
\end{equation}
where $B_i = \{j: j_N \in [1, N], j_C \in [i_C], j_L \in [1, L]\}$,
and $i = (i_N, i_C, i_L)$ indexing the activations $x \in \mathbb R^{N\times C \times L}$.\newline

\textbf{Layer Normalization (LN):} Alleviating inter-dependency among batch-normalized activations, LN computes mean and variance using input from individual layer neurons. LN benefits recurrent networks but may face challenges with convolutional layers due to spatial variations in visual data. Formulated as Equation~\ref{general_transform} with $B_i = \{j: j_N \in [i_N], j_C \in [1, C], j_L \in [1, L]\}$.\newline

\indent \textbf{Instance Normalization (IN):} This approach normalizes each sample individually, particularly aimed at eliminating image style information. IN enhances specific deep neural networks (DNNs) by computing mean values and standard deviations in the spatial domain. Applied as Equation~\ref{general_transform} with $B_i = \{j: j_N \in [i_N], j_C \in [i_C], j_L \in [1, L]\}$.\newline

\textbf{Group Normalization (GN):} GN partitions neurons into groups and autonomously standardizes layer inputs for each sample within these groups. In visual tasks with small batch sizes, GN is useful for applications like object detection and segmentation. GN calculates statistics along the $L$ axis, restricting computation to subgroups of channels. Equivalent to IN when $G=C$ and identical to LN when $G=1$.\newline

\textbf{Mixture Normalization (MN):} In deep neural networks (DNNs), neuron activations exhibit multiple modes of variation due to non-linearities, challenging the Gaussian distribution assumption of batch normalization (BN)~\cite{ioffe2015batch}. Mixture Normalization (MN)~\cite{kalayeh2019training} approaches BN from a Fisher kernels perspective, using a Gaussian Mixture Model (GMM) to normalize based on multiple means and standard deviations associated with the clusters. In convolutional neural networks, MN outperforms BN in terms of convergence and accuracy in supervised learning tasks. Implemented in two stages: estimation of GMM parameters using the Expectation-Maximization (EM) algorithm~\cite{Dempster77maximumlikelihood}, followed by neuron activations normalization and aggregation using posterior probabilities during deep neural network training.\newline
\indent The distribution $p_\theta$ characterizing the data is represented as a parameterized Gaussian Mixture Model (GMM). Let $x \in \mathbb{R}^D$, and $\theta = \{\lambda_k, \mu_k, \Sigma_k: k = 1, ..., K\}$,
\begin{equation}
\label{gmm}
p(x) = \sum_{k=1}^K \lambda_kp(x|k),\ \text{s.t.}\ \forall_k\ :\ \lambda_k\ \ge 0,\ \sum_{k=1}^{K}\lambda_k=1,
\end{equation}
where
\begin{equation}
p(x|k) = \frac{1}{(2\pi)^{D/2}\lvert\Sigma_k\rvert^{1/2}}\exp\left(-\frac{( x-\mu_k)^T\Sigma_k^{-1}( x-\mu_k)}{2}\right),
\end{equation}
represents the $k^{th}$ Gaussian in the mixture model $p(x)$, with $\mu_k$ as the mean vector and $\Sigma_k$ as the covariance matrix.\newline
The posterior probability that $x$ is generated by the $k^{th}$  component in the mixture model is defined as:
\begin{equation}
\tau_k(x) = p(k|x) = \frac{\lambda_k p(x|k)}{\sum_{j=1}^K\lambda_j p(x|j)},
\end{equation}
Based on these assumptions and the general transform in Equation~\eqref{general_transform}, the Mixture Normalizing Transform for a given $x_i$ is defined as
\begin{equation}
\label{mn_aggregation}
\hat{x}_{i} = \sum_{k=1}^K \frac{\tau_k(x_{i})}{\sqrt{\lambda_k}}\hat{x}_{i}^k,
\end{equation}
given
\begin{equation}
\label{mn_norm}
\hat{x}_{i}^k = \frac{v_{i}^k}{\sqrt{\mathbb E_{B_i}[\hat{\tau}_k(x).(v^k)^2]+\epsilon}},
\end{equation}
where
\begin{equation*}
\hat{\tau}_k(x_{i}) = \frac{\tau_k(x_{i})}{\sum_{j \in B_i} \tau_k(x_{j})},
\end{equation*}
and
\begin{equation*}
v_{i}^k=x_{i} - \mathbb E_{B_i}[\hat{\tau}_k(x).x], 
\end{equation*}
represents the normalized contribution of $x_i$ with respect to the mini-batch $B_i = \{j: j_N \in [i_N], j_C \in [i_C], j_L \in [1, L]\}$ in estimating the statistical measures of the $k^{th}$ Gaussian component.\newline
%

\section{Contribution: Unsupervised Adaptive Normalization }
Mixture Normalization (MN) effectively stabilizes deep neural network training, but it encounters limitations in its two-stage process, which involves the Expectation-Maximization (EM) algorithm, introducing complexities. The static parameters of MN do not adapt to the changing distributions of neuron activations during learning and weight updates. Recognizing the need for a more adaptive and streamlined approach, we propose Unsupervised Adaptive Normalization (UAN). In response to these challenges, UAN operates as a one-stage algorithm, embodying an "unsupervised" approach where prior knowledge of mixture components is absent. In this context, "clusters" refers to a mixture component within the latent space modeled as a Gaussian mixture. UAN estimates mixture component parameters as network weights, dynamically adjusting to the target task and evolving alongside deep neural network weights. This unique one-stage algorithm achieves mixture parameter estimation and target task resolution during a unified training process. Unlike MN, where normalization parameters remain constant, UAN updates these parameters throughout learning, ensuring the formation of adaptive clusters in line with the evolving distribution of neuron activations.

In implementing this approach, we assume the presence of a latent space where clusters are modeled as a mixture model. Specifically, we consider a Gaussian mixture model with parameters $\theta = \{\lambda_k, \mu_k, \Sigma_k: k = 1, ..., K\}$, where $K$ denotes the total number of defined clusters. It is important to note that the assumption of a Gaussian distribution is not mandatory; alternative distributions can be explored. In such cases, the corresponding parameters are estimated accordingly. The normalization of each neuron activation takes place using Equation~\ref{mn_aggregation}, where all clusters contribute to the normalization process for each activation.
\begin{figure}[!htbp]
\centering
\includegraphics[width=0.45\textwidth]{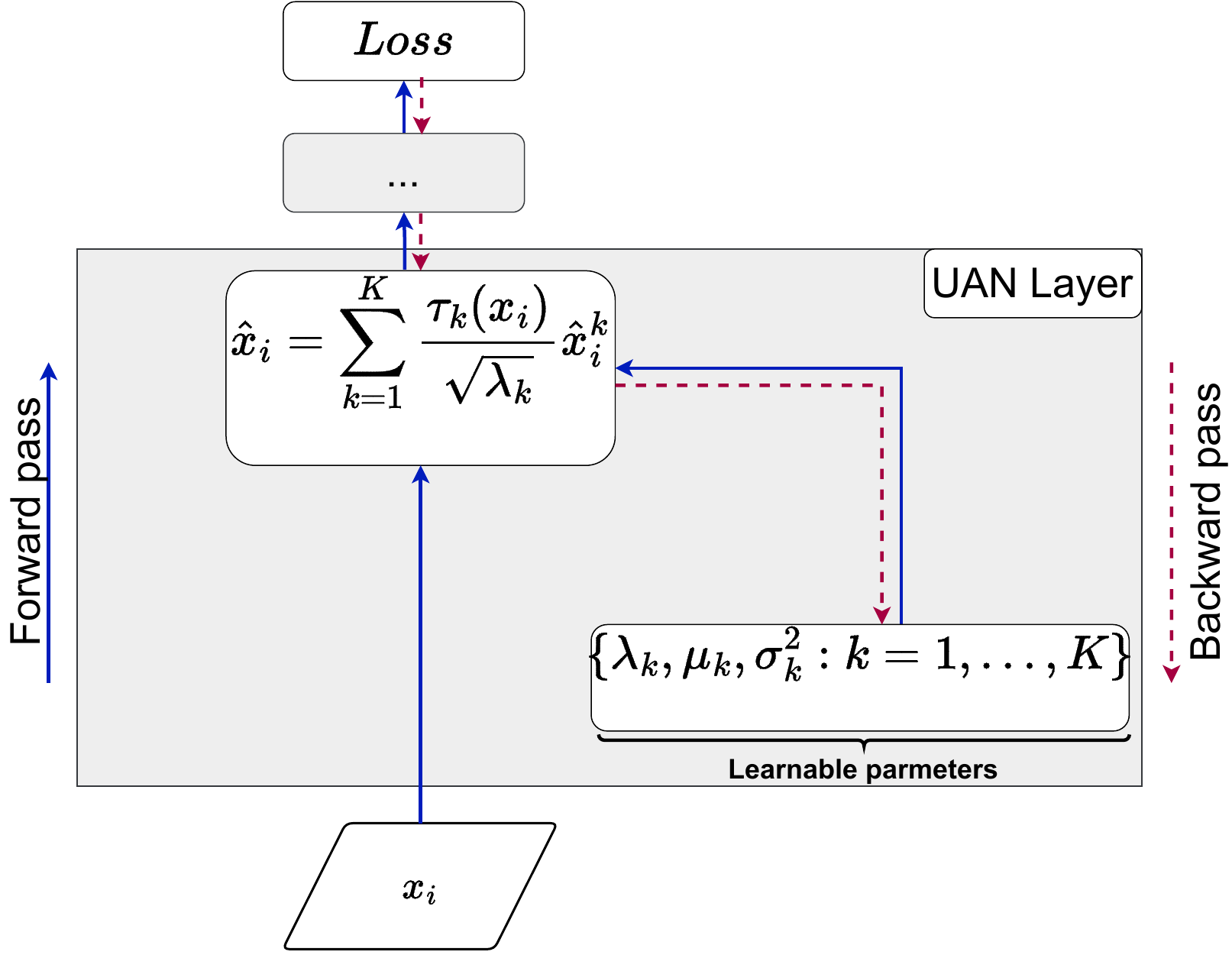}
\caption{Unsupervised Adaptive Normalization (UAN) applied on neuron activation $x_i$. The parameter $K$ corresponds to the number of clusters estimated during the training process. In the forward pass, the $K$ cluster parameters $\{\lambda_k, \mu_k, \sigma_k^2\}$ are used to normalize the activation $x_i$ through Equation~\ref{mn_aggregation}. The resulting normalized activation serves as input for the Deep Neural Network (DNN). After loss computation, in the backward pass, the parameters $\{\lambda_k, \mu_k, \sigma_k: k = 1, ..., T\}$ undergo updates based on the target task.}
\label{fig:cn_layer}
\end{figure}

Throughout the training of the deep neural network,  cluster parameters are set at the beginning to fulfill the conditions $\sum_{i=1}^K \lambda_k = 1$ and $\sigma_k > 0$, where these parameters act as weights. To adapt these cluster parameters, we use backpropagation and a moving average technique within mini-batches (Figure~\ref{fig:cn_layer} and Algorithm~\ref{alg:two}). The former aligns neuron activations with the target task, dynamically evolving parameters alongside deep neural network weights. Simultaneously, the latter serves as an approximation of the EM approach tailored to neuron activations. This one-stage process in UAN allows the simultaneous estimation of mixture component parameters and resolution of the target task in deep neural networks. Unlike methods with static parameters that struggle to adapt during learning, UAN demonstrates adaptability and efficiency in overcoming evolving neuron activation distributions and weight updates.

\begin{algorithm}[!htbp]
\caption{ Unsupervised Adaptive Normalization Layer}\label{alg:two}
\SetKwInOut{KwIn}{Input}
\SetKwInOut{KwOut}{Output}

\KwIn{Deep neural network $Net$ with trainable parameters $\Theta$; subset of activations $\{x_i\}_{i=1}^m$; number of clusters $K$; momentum $m$; $update \in \{weight, moving\ average\}$
}
\KwOut{Unsupervised Adaptive-Normalized deep neural network for inference, $Net^{inf}_{UAN}$}
    $Net^{tr}_{UAN} = Net$ // {\small \it Training UAN deep neural network }\\ 
    Initialize the parameters for each cluster as follows: \newline
    ($\lambda_k$, $\mu_k$, $\sigma^2_k$) for $k \in \{1, ..., K\}$, subject to the conditions that
    $\sum_{k=1}^K \lambda_k = 1$ and $\sigma_k > 0$ for all $k \in \{1, ..., K\}$. 
    
    \For{$i \gets 1$ to $m$}{
        \begin{itemize}
            \item Add transformation $\hat{x}_i$ = $UAN$($x_i$) to $Net^{tr}_{UAN}$ 
            using Equation~\ref{mn_aggregation}
            \item Modify each layer in $Net^{tr}_{UAN}$ with input $x_i$\\ to take $\hat{x}_i$ instead
        \end{itemize}
    }
    \If{$update = weight$}{ // {\small \it Update parameters as weights of the neural network}\\
    Train $Net^{tr}_{UAN}$ to optimize the parameters: $\Theta = \Theta \cup \{\lambda_k, \mu_k, \sigma^2_k\}_{k=1}^K$ \\
    }
    \If{$update = moving\ average$}{ // {\small \it Update parameters using moving average according the batch} \\\
    \For{$k \gets 1$ to $K$}{
        $\lambda_k \gets m\times \lambda_k + (1-m)\times mean(\tau_k)$ \\
        $\mu_k \gets m\times \mu_k + (1 - m) \times mean(\hat{x}^k)$ \\
        $\sigma^2_k \gets m\times \sigma^2_k + (1 - m)\times variance(\hat{x}^k)$
        
    } 
    }
    $Net^{inf}_{UAN} = Net^{tr}_{UAN}$ // {\small \it Inference UAN network with frozen parameters}\\ 
    \For{$i \gets 1$ to $m$}{
            \begin{itemize}
                \item Transform $\hat{x}_i = UAN(x_i)$ using Equation~\ref{mn_aggregation}
                \item $\hat{x}_{i} = \sum_{k=1}^K \frac{\tau_k(x_{i})}{\sqrt{\lambda_k}}(\frac{x_i - \mu_k}{\sigma_k})$
                \item Modify each layer in $Net^{inf}_{UAN}$ with input $x_i$\\ to take $\hat{x}_i$ instead
            \end{itemize}
    }
\end{algorithm}

\section{Experiments}
In the upcoming experiments, our goal is to evaluate the performance of the Unsupervised Adaptive Normalization (UAN) method and compare it with Batch Normalization (BN) and Mixture Normalization (MN). We suggest applying UAN to tasks involving classification (Section~\ref{section:experiment1}), 
and domain adaptation (Section~\ref{section:experiment3}).\newline
UAN, a deep neural network layer equipped with Gaussian mixture parameters, undergoes a distinctive initialization process during training. The mean ($\mu$) parameters are set with random values uniformly distributed between $-1.0$ and $1.0$, establishing a varied starting point conducive to capturing central tendencies. The standard deviation ($\sigma$) parameters receive initialization with strictly positive random values within the range of 0.001 to 0.01, ensuring numerical stability and facilitating the exploration of diverse data patterns. Additionally, the mixing coefficients ($\lambda$) parameters are initialized with values spanning $0.01$ to $0.99$. These values are normalized to ensure their collective sum equals $1$, thereby establishing a well-balanced and probabilistically sound foundation for incorporating prior knowledge within the Gaussian mixture model.
\subsection{Datasets}
\label{datasets}
The experiments in this study utilize several benchmark datasets widely recognized within the classification community:
\begin{itemize}
\item \textbf{CIFAR-10:} A dataset with 50000 training images and 10000 test images, each of size $32\times32$ pixels, distributed across 10 classes~\cite{cifar10_datasets}.
\item \textbf{CIFAR-100:} Derived from the Tiny Images dataset, it consists of 
50000 training images and 10000 test images of size $32\times32$, divided into 100 classes grouped into 20 superclasses~\cite{cifar100_datasets}.
\item \textbf{Tiny ImageNet:} A reduced version of the ImageNet dataset, containing 200 classes with 500 training images and 50 test images per class~\cite{le2015tiny}.
\item \textbf{MNIST digits:} Contains 70000 grayscale images of size $28\times28$ representing the 10 digits, with around 6000 training images and 1000 testing images per class~\cite{mnist_datasets}.
\item \textbf{SVHN:} A challenging dataset with over 600000 digit images, focusing on recognizing digits and numbers in natural scene images~\cite{sermanet2012convolutional}.
\end{itemize}

\subsection{Unsupervised Adaptive Normalization on Classification task}
\label{section:experiment1}
\subsubsection{Shallow Convolutional Neural Network}
In this experimental setup, we utilize a Shallow Convolutional Neural Network (Shallow CNN)~\cite{lecun1998gradient} architecture following the design principles outlined in the Mixture Normalization (MN) paper. The Shallow CNN comprises four convolutional layers, each activated by a Rectified Linear Unit (ReLU)~\cite{relu}, and subsequently normalized using a Batch Normalization (BN) layer. One notable challenge with BN arises when employing non-linear functions (e.g., ReLU) post-activation normalization. To address this issue, we introduce Unsupervised Adaptive Normalization (UAN) in the Shallow CNN, replacing the BN layer. We demonstrate how UAN resolves this issue, improving convergence and overall performance during training.

During the training phase on CIFAR-10, CIFAR-100, and Tiny ImageNet datasets, we apply the MN method, which involves estimating a Gaussian mixture model through Maximum Likelihood Estimation (MLE)~\cite{bishop2006pattern}. We utilize three components identified by the Expectation-Maximization (EM) algorithm~\cite{Dempster77maximumlikelihood} for MN. For a fair comparison, we also employ $K=3$ clusters for UAN. We conduct experiments with varying learning rates from 0.001 to 0.005, use a batch size of 256, and train for 100 epochs with the AdamW optimizer~\cite{loshchilov2017fixing}.
To assess the efficacy of normalization methods, we substitute the third BN layer in Shallow CNN with an MN layer and replicate this process for the UAN layer. 

\begin{figure*}[htbp]
    \centering
    \includegraphics[width=1\textwidth, height=4.5cm]{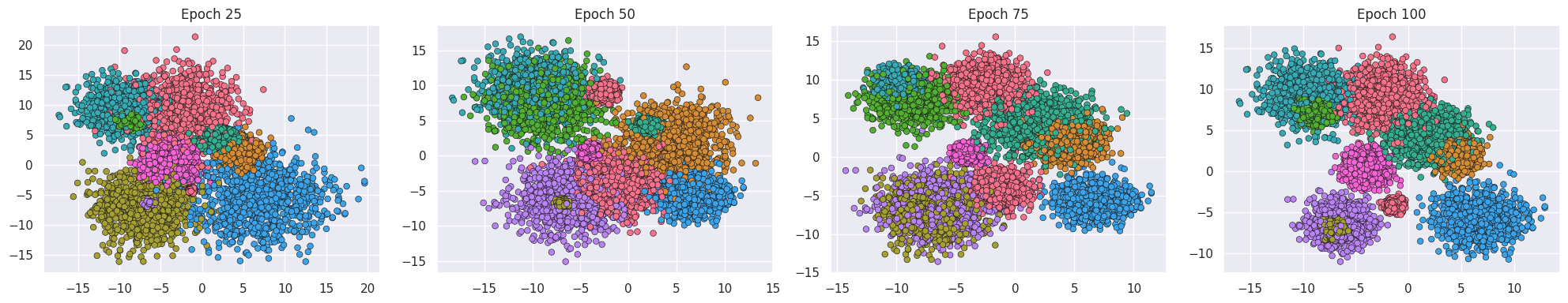}
    \caption{t-SNE visualization of activations in latent space after 25, 50, 70 and 100  training epochs. Unsupervised Adaptive Normalization (UAN) on CIFAR-10, showing the formation and refinement of class-specific clusters over training epochs.}
    \label{fig:cluster1}
\end{figure*}

During the training process on CIFAR-10, we captured periodic snapshots every 25 epochs while utilizing the UAN normalization. Subsequently, the model was deployed on a random CIFAR-10 batch, and the resulting activations underwent t-SNE visualization. The visualizations unveiled the formation of clusters aligning with the target classes predicted by the model. We observe an improvement in this clustering as the training, contributes to a comprehensive enhancement in performance, as depicted in Figure~\ref{fig:cluster1}.

\begin{figure*}[htbp]
    \centering
    \begin{subfigure}[b]{0.25\textwidth}
        \centering
        \includegraphics[width=\textwidth]{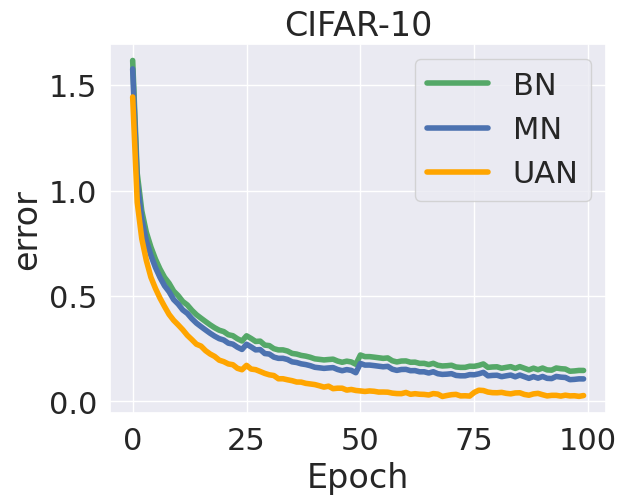}
        \caption{CIFAR-10}
        \label{fig:bn}
    \end{subfigure}
    \hfill
    \begin{subfigure}[b]{0.25\textwidth}
        \centering
        \includegraphics[width=\textwidth]{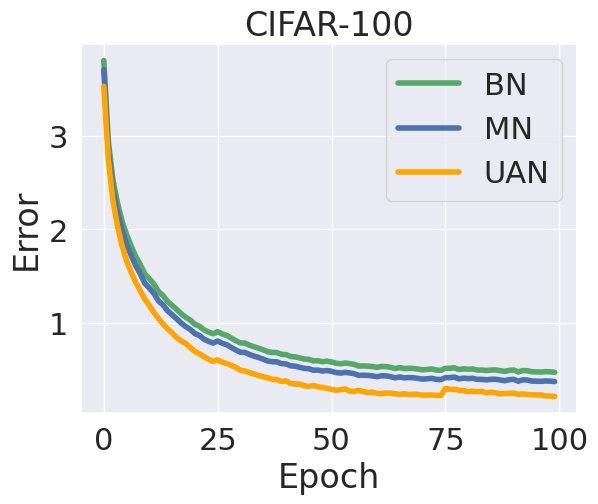}
        \caption{CIFAR-100}
        \label{fig:CN-Patches}
    \end{subfigure}
    \hfill
    \begin{subfigure}[b]{0.25\textwidth}
        \centering
        \includegraphics[width=\textwidth]{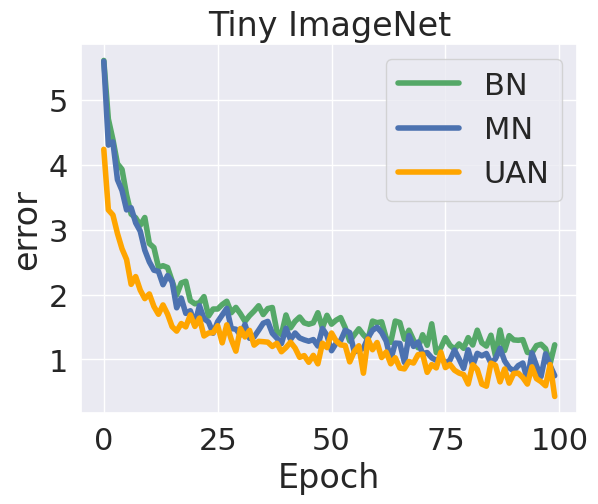}
        \caption{Tiny ImageNet}
        \label{fig:CN-Channels}
    \end{subfigure}
    \caption*{Learning rate = 0.001}
    \label{fig:cifar100_loss}


    \begin{subfigure}[b]{0.25\textwidth}
        \centering
        \includegraphics[width=\textwidth]{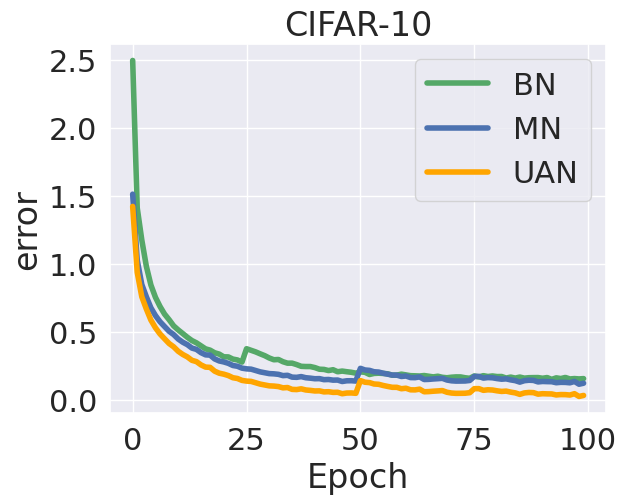}
        \caption{CIFAR-10}
        \label{fig:figure1}
    \end{subfigure}
    \hfill
    \begin{subfigure}[b]{0.25\textwidth}
        \centering
        \includegraphics[width=\textwidth]{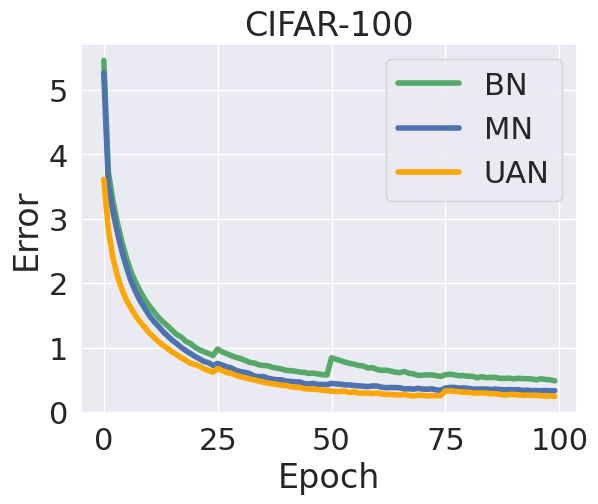}
        \caption{CIFAR-100}
        \label{fig:figure2}
    \end{subfigure}
    \hfill
    \begin{subfigure}[b]{0.25\textwidth}
        \centering
        \includegraphics[width=\textwidth]{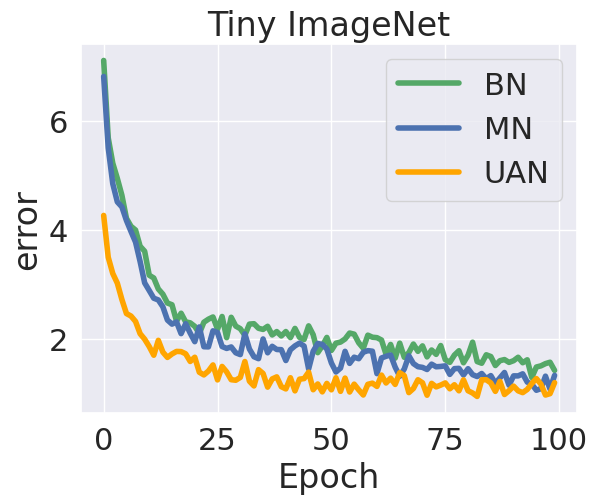}
        \caption{Tiny ImageNet}
        \label{fig:figure3}
    \end{subfigure}
    \caption*{Learning rate = 0.005}
    \label{fig:second_set_of_figures}
    \caption{Test error curves when the Shallow CNN architecture is trained under different learning rates.}
    \label{fig:all_figures}
\end{figure*}

Figure~\ref{fig:all_figures} illustrates that UAN outperforms both BN and MN in terms of convergence speed. This faster convergence is evident in the improved performance on the validation dataset. Specifically, UAN shows an average accuracy increase of \textbf{2\%} on CIFAR-10, \textbf{3\%} on CIFAR-100, and \textbf{4\%} on Tiny ImageNet. These enhancements are consistent across different class numbers (10, 100, 200) and are even more pronounced when the learning rate is raised from 0.001 to 0.005. The increasing disparity in convergence rates at higher learning rates highlights the superiority of our normalization method in effectively utilizing higher learning rates to achieve better training results. These findings align with the observed improvement in cluster formation, affirming the positive impact of the proposed normalization method on model performance.
\newline
\indent To validate the hypothesis that Unsupervised Adaptive Normalization (UAN) substantially aids in accelerating the convergence of neural network training, we  conduct the upcoming experiment using a Deep Convolutional Neural Network on the CIFAR-100 dataset.
\subsubsection{Deep Convolutional Neural Network}
we have shown that integrating UAN accelerates convergence and enhances performance in model using batch normalization layer as baseline. Experiments were conducted on a shallow architecture with four convolutional layers, prompting the question of whether this behavior extends to more complex architectures. To investigate that, we choose DenseNet~\cite{huang2017densely}, leveraging 40 and 100-layer architectures. All models undergo 200 epochs on CIFAR-100, with a batch size of 64, Nesterov’s accelerated gradient~\cite{bengio2013advances}, and initial learning rate of $0.1$, reduced by a factor of 10 at $50\%$ and $75\%$ of training epochs. Weight decay and momentum are $10^{-4}$ and $0.9$, respectively.\newline 
\indent To provide a comparative viewpoint, we substitute the Batch Normalization (BN) layer in both the DensetNet-40 and DenseNet-100 architectures with layers such as MN and UAN, resulting in separate models. In the case of the MN layer, we employ $5$ Gaussian components. As in the initial experiment, for a fair comparison, we designate MN components as individual clusters ($K=5$) for models with UAN layer.\newline
\begin{figure*}[htbp]
    \centering
    \begin{subfigure}[b]{0.22\textwidth}
        \centering
        \includegraphics[width=\textwidth]{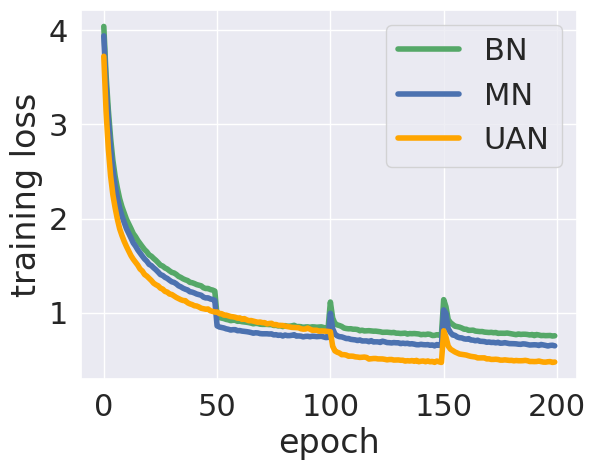}
        \caption{}
        \label{fig:bn}
    \end{subfigure}
    \hfill
    \begin{subfigure}[b]{0.22\textwidth}
        \centering
        \includegraphics[width=\textwidth]{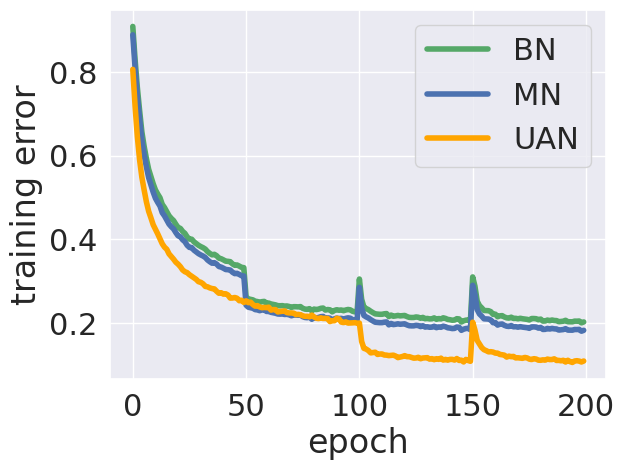}
        \caption{}
        \label{fig:CN-Patches}
    \end{subfigure}
    \hfill
    \begin{subfigure}[b]{0.22\textwidth}
        \centering
        \includegraphics[width=\textwidth]{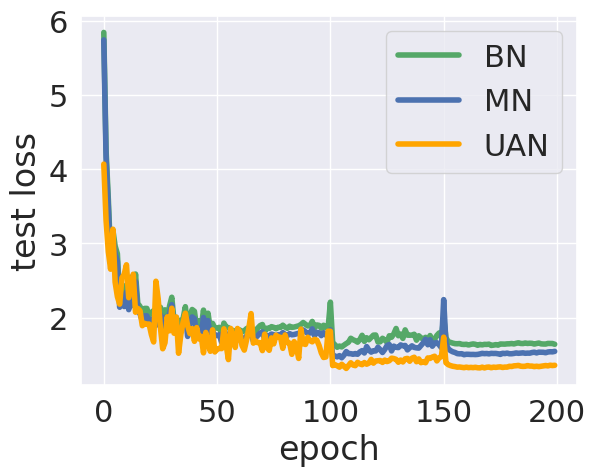}
        \caption{}
        \label{fig:CN-Channels}
    \end{subfigure}
    \hfill
    \begin{subfigure}[b]{0.22\textwidth}
        \centering
        \includegraphics[width=\textwidth]{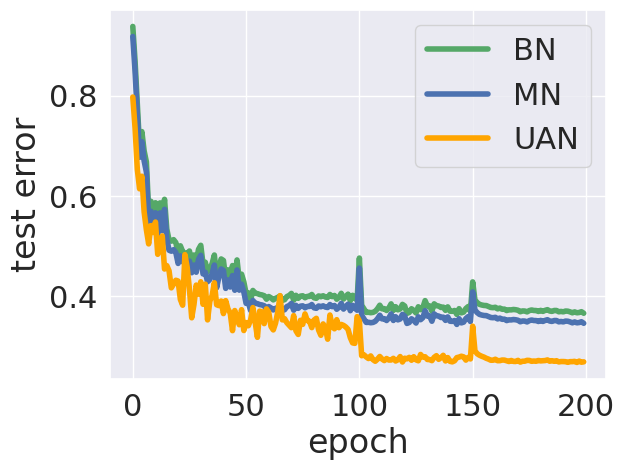}
        \caption{}
        \label{fig:bn}
    \end{subfigure}
    \caption*{DenseNet-40}
    \label{fig:densenet40}

    \begin{subfigure}[b]{0.22\textwidth}
        \centering
        \includegraphics[width=\textwidth]{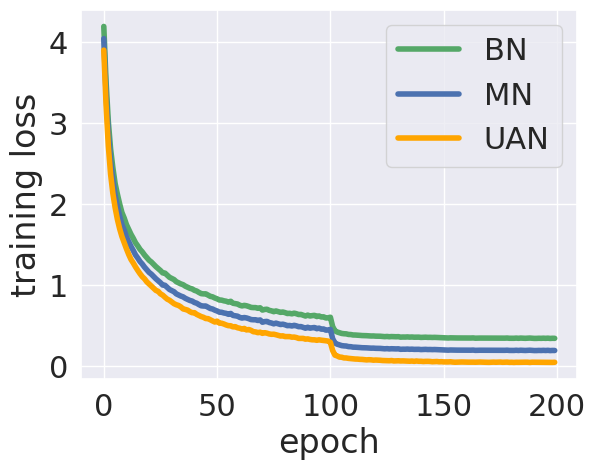}
        \caption{}
    \end{subfigure}
    \hfill
    \begin{subfigure}[b]{0.22\textwidth}
        \centering
        \includegraphics[width=\textwidth]{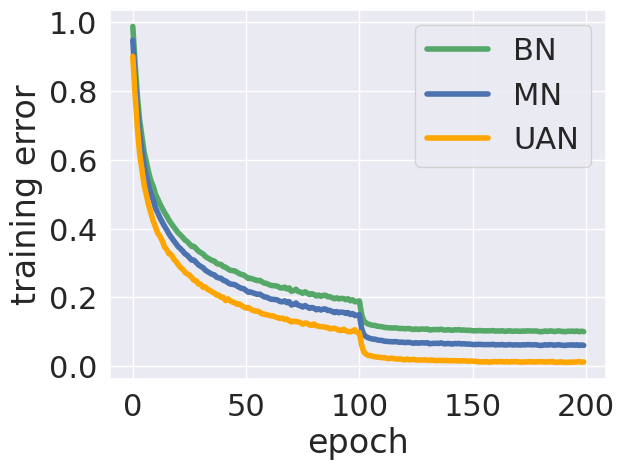}
        \caption{}
        \label{fig:figure2}
    \end{subfigure}
    \hfill
    \begin{subfigure}[b]{0.22\textwidth}
        \centering
        \includegraphics[width=\textwidth]{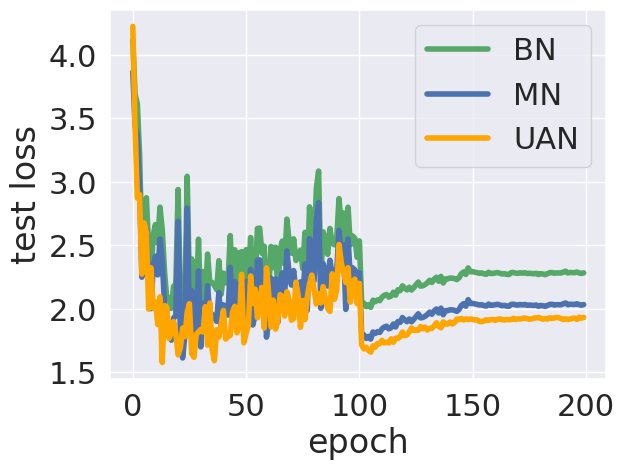}
        \caption{}
        \label{fig:figure3}
    \end{subfigure}
    \hfill
    \begin{subfigure}[b]{0.22\textwidth}
        \centering
        \includegraphics[width=\textwidth]{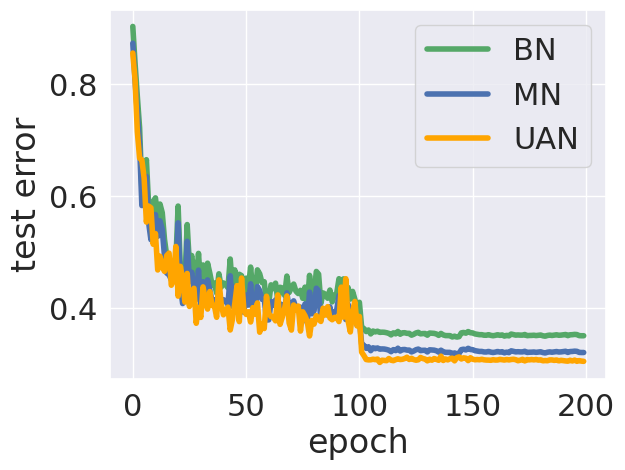}
        \caption{}
    \end{subfigure}
    \hfill
    \caption*{DenseNet-100}
    \label{fig:densenet100}
    \caption{
    Figures present CIFAR-100 experiments with DenseNet~\cite{huang2017densely}. The graphs showcase training/test errors and cross-entropy loss for DenseNet-40 (40 layers) and DenseNet-100 (100 layers). Unsurprisingly, introducing Unsupervised Adaptive Normalization remarkably accelerates the training process, facilitating optimization. Simultaneously, it exhibits superior generalization, consistently outperforming batch normalization and mixture normalization counterparts in the error curve.}
    \label{fig:densenet}
\end{figure*}
\indent Figure~\ref{fig:densenet}  illustrates that UAN offers more than just accelerated training optimization—it consistently demonstrates superior generalization across various architectural setups. This advantage is particularly evident in the complex DenseNet architecture. Using DenseNet-40 on CIFAR-100, the test error for Batch Normalization (BN) and Mixture Normalization (MN) is respectively  \textbf{0.38} and \textbf{0.37}, while the UAN achieves a notably lower test error of \textbf{0.36}. With the deeper DenseNet-100, the test errors for BN and MN drop respectively to \textbf{0.3} and \textbf{0.29}, while the UAN variant achieves the best test error of \textbf{0.28}.

\subsection{Unsupervised Adaptive Normalization on Domain Adaptation}
\label{section:experiment3}

In this experimental setup, we demonstrate how the effectiveness of Unsupervised Adaptive Normalization enhancing local representations can lead to significant advancements in domain adaptation. As outlined in~\cite{farahani2021brief}, domain adaptation involves leveraging knowledge gained by a model from a related domain, where there is an abundant amount of labeled data, to improve the model's performance in a target domain with limited labeled data. We consider two domains ($K=2$) as two clusters: the "source domain" and the "target domain" 
in conjunction with AdaMatch~\cite{berthelot2021adamatch}. This method combines the tasks of unsupervised domain adaptation (UDA), semi-supervised learning (SSL), and semi-supervised domain adaptation (SSDA). In UDA, we have access to a labeled dataset from the source domain and an unlabeled dataset from the target domain. The goal is to train a model capable of effectively generalizing to the target dataset. It's essential to note that the source and target datasets exhibit variations in distribution. Specifically, we use the MNIST dataset as the source dataset (cluster $k=1$), while the target dataset is SVHN (cluster $k=2$). Both datasets encompass various factors of variation, including texture, viewpoint, appearance, etc., and their domains, or distributions, are distinct.\newline
\begin{figure*}[!t]
  \centering
  \begin{subfigure}{0.65\textwidth}
    \includegraphics[width=\linewidth]{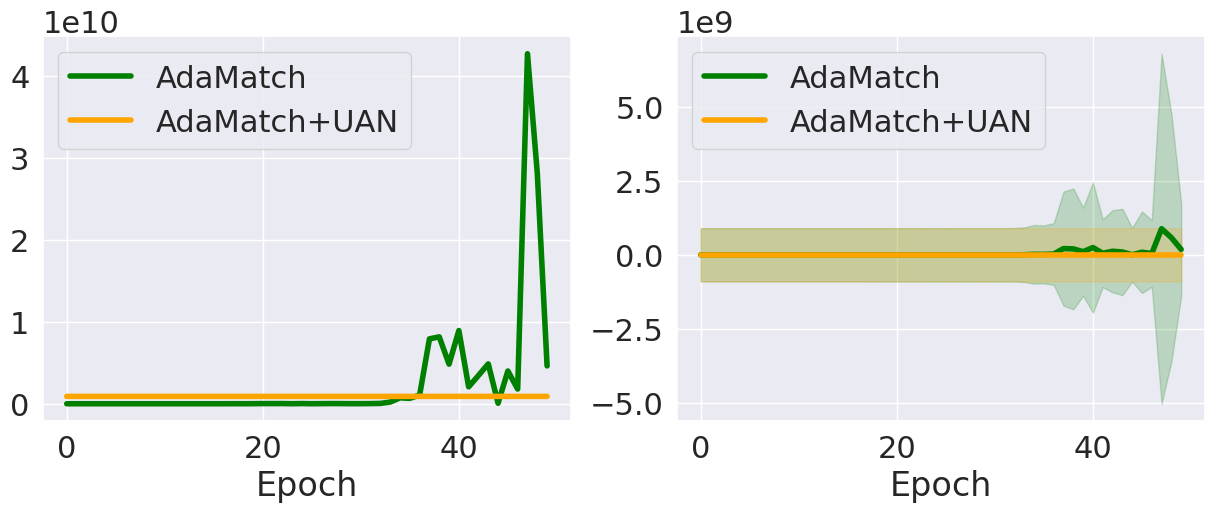}
    \caption{Source domain (MNIST)}
  \end{subfigure}
  \begin{subfigure}{0.65\textwidth}
    \includegraphics[width=\linewidth]{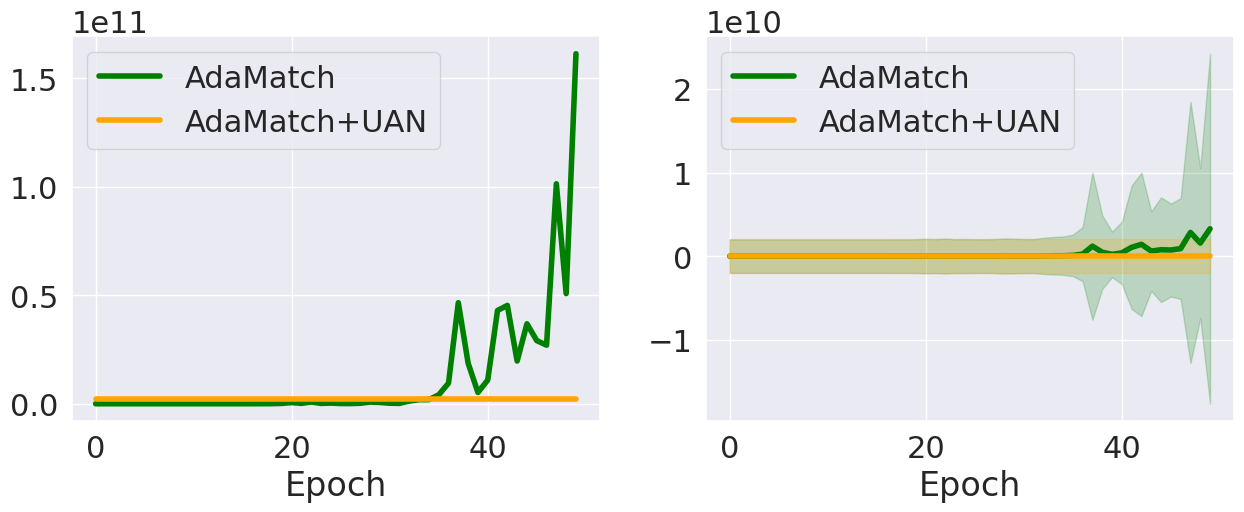}
    \caption{Target domain (SVHN)}
  \end{subfigure}
  \caption{Evolution of the gradient variance during the training of AdaMatch and AdaMatch+UAN models on the source (MNIST) and target (SVHN) domains. The figures on the left correspond to the maximum gradient variance for each epoch, while the figures on the right correspond to the average gradient variance per epoch.}
  \label{fig:adamatch_gradient}
\end{figure*}
\indent AdaMatch~\cite{keras_adamatch} is trained from scratch using wide residual networks~\cite{zagoruyko2016wide} on dataset pairs, forming the baseline model. The model training involves employing the Adam optimizer~\cite{kingma2014adam} with a cosine decay schedule, gradually reducing the initial learning rate initialized at 0.03. In this practical scenario, Mixture Normalization cannot be applied because we aim to tailor normalization according to domains, which may not necessarily constitute a Gaussian mixture. UAN serves as the initial layer in AdaMatch, resulting in the creation of AdaMatch model with Unsupervised Adaptive Normalization(AdaMatch+UAN). This incorporation enables the cluster identifier (source domain and target domain) to influence the image normalization process.\newline
\begin{table}[t]
    \centering
    \begin{tabular}{llllll}
        \hline
        \multicolumn{5}{c}{\textbf{MNIST (source domain)}}\\
        \hline
        model  & accuracy & precision & recall & f1-score  \\
        \hline
        AdaMatch & 97.36 & 87.33 & 79.39 & 78.09 \\
        \hline
        AdaMatch+UAN & 98.9 & 98.5 & 98.90 & 98.95 \\
        \hline
    \end{tabular}

    \begin{tabular}{llllll}
        \hline
        \multicolumn{5}{c}{\textbf{SVHN (target domain)}}\\
        \hline
        model  & accuracy & precision & recall & f1-score  \\
        \hline
        AdaMatch & 25.08 & 31.64 & 20.46 & 24.73 \\
        \hline
        AdaMatch+UAN & 33.4 & 43.83 & 40.28 & 42.87 \\
        \hline
    \end{tabular}
    \caption{Comparing model performance: AdaMatch vs. AdaMatch+UAN on MNIST (source) and SVHN (target) datasets.}
    \label{table:adamatch}
\end{table}
\indent 
Table~\ref{table:adamatch} unmistakably underscores a noteworthy improvement in validation metrics resulting from the integration of Unsupervised Adaptive Normalization. This progress is evident, translating into a substantial accuracy boost of \textbf{8.32\%}. This notable enhancement significantly strengthens the performance of the AdaMatch model, contributing to a markedly accelerated convergence during training.

Figure~\ref{fig:adamatch_gradient}  highlights the invaluable role of Unsupervised Adaptive Normalization in stabilizing the gradient throughout the training process, providing advantages for both the source and target domains. This stabilization significantly contributes to accelerated convergence and an overall improvement in model performance.\newline
\section{Computational Complexity}
\label{section:computational}
Compared to Batch Normalization~\cite{ioffe2015batch} and Mixture Normalization~\cite{kalayeh2019training}, the computational overhead of adaptative normalization arises from estimating cluster parameters. Let $B$ represent a batch of activations in $\mathbb{R}^d$, where $d$ is the dimensionality of activations.\newline
\indent For Batch Normalization, mean ($\mu_B$) and standard deviation ($\sigma_B$) are computed considering the entire batch $B$, leading to a computational complexity of $\mathcal{O}(d)$. Learning parameters $\gamma$ and $\beta$ for each batch dimension adds a parameter learning complexity of $\mathcal{O}(d)$. Thus, the overall complexity, accounting for both computational and parameter aspects, is $\mathcal{O}(d)$ for the entire batch $B$.\newline
\indent Mixture normalization faces computational challenges in estimating Gaussian mixture model (GMM) parameters accurately. Using K-means++ for seeding, complexity for batch $B$ with dimension $d$ is $\mathcal{O}(d \cdot K \cdot \log(K))$. In the EM algorithm, per-iteration complexity is $\mathcal{O}(d \cdot n \cdot K)$, mainly determined by initialization, also $\mathcal{O}(d \cdot K \cdot \log(K))$. Like Batch Normalization, mean ($\mu_k$) and standard deviation ($\sigma_k$) are computed per cluster $k$, $\mathcal{O}(K \cdot d)$. Parameter learning complexity ($\gamma$, $\beta$) per batch dimension is $\mathcal{O}(d)$. Overall complexity, considering both computational and parameter aspects, is $\mathcal{O}(d \cdot n \cdot K)$ for batch $B$.\newline
\begin{table*}[!t]
    \centering
    \begin{tabular}{|l|c|c|}
        \hline
        \textbf{Normalization Method} & \textbf{Computational Complexity} & \textbf{Parameter Complexity} \\
        \hline
        Batch Normalization (BN) & $\mathcal{O}(d)$ & $\mathcal{O}(d)$ \\
        \hline
        Mixture Normalization (MN) & $\mathcal{O}(d \cdot n \cdot K)$ & $\mathcal{O}(d)$ \\
        \hline
        Unsupervised Adaptive Normalization (UAN) & $0$ & $\mathcal{O}(K \cdot d)$ \\
        \hline
    \end{tabular}
    \caption{Complexities in Estimating Parameters for Normalization Methods. Computational Complexity corresponds to the time required for parameter estimation, while Parameter Complexity corresponds to the number of parameters to be estimated. Here, $d$ represents the dimension of the data, $K$ denotes the number of cluster centers, and $n$ the number of samples.}
    \label{tab:normalization_complexity}
\end{table*}
\indent Unsupervised Adaptive Normalization (UAN) estimates three learnable parameters for each cluster $k$: $\lambda_k$ of dimension 1, and $\mu_k$ and $\sigma_k$ of dimension $d$. This results in $2 \times K \times d + K$ parameters to be learned, with a complexity of $\mathcal{O}(K \cdot d)$. Unlike MN, UAN is performed in a single-stage process, and the complexity is solely associated with parameter estimation during backpropagation. A comparison is presented in Table~\ref{tab:normalization_complexity}.

\section{Conclusion}

In this paper, we introduced Unsupervised Adaptive Normalization (UAN), a novel one-stage unsupervised normalization technique for deep neural networks. UAN provides mixture components during the training of deep neural networks, tailoring them specifically for the target task. This method is efficient as it computes cluster parameters simultaneously with the neural network's weight adjustments, thus minimizing computational overhead compared to multi-stage normalization approaches or classical ones. UAN stabilizes neural network training across a range of applications, including supervised classification, 
and domain adaptation, proving its versatility and effectiveness in various contexts. In future work, we plan to further validate the efficacy of our approach on different context as Generative Adversarial Networks (GANs) and different types of data, particularly focusing on its adaptability to architectures that handle multimodal data.\newline

\bibliographystyle{IEEEtran}
\bibliography{sample-base}

\begin{thebibliography}{10}
\providecommand{\url}[1]{#1}
\csname url@samestyle\endcsname
\providecommand{\newblock}{\relax}
\providecommand{\bibinfo}[2]{#2}
\providecommand{\BIBentrySTDinterwordspacing}{\spaceskip=0pt\relax}
\providecommand{\BIBentryALTinterwordstretchfactor}{4}
\providecommand{\BIBentryALTinterwordspacing}{\spaceskip=\fontdimen2\font plus
\BIBentryALTinterwordstretchfactor\fontdimen3\font minus \fontdimen4\font\relax}
\providecommand{\BIBforeignlanguage}[2]{{%
\expandafter\ifx\csname l@#1\endcsname\relax
\typeout{** WARNING: IEEEtran.bst: No hyphenation pattern has been}%
\typeout{** loaded for the language `#1'. Using the pattern for}%
\typeout{** the default language instead.}%
\else
\language=\csname l@#1\endcsname
\fi
#2}}
\providecommand{\BIBdecl}{\relax}
\BIBdecl

\bibitem{lecun2002efficient}
Y.~LeCun, L.~Bottou, G.~B. Orr, and K.-R. M{\"u}ller, ``Efficient backprop,'' in \emph{Neural networks: Tricks of the trade}.\hskip 1em plus 0.5em minus 0.4em\relax Springer, 2002, pp. 9--50.

\bibitem{ioffe2015batch}
S.~Ioffe and C.~Szegedy, ``Batch normalization: Accelerating deep network training by reducing internal covariate shift,'' in \emph{International conference on machine learning}.\hskip 1em plus 0.5em minus 0.4em\relax PMLR, 2015, pp. 448--456.

\bibitem{ba2016layer}
J.~L. Ba, J.~R. Kiros, and G.~E. Hinton, ``Layer normalization,'' \emph{arXiv preprint arXiv:1607.06450}, 2016.

\bibitem{ulyanov2016instance}
D.~Ulyanov, A.~Vedaldi, and V.~Lempitsky, ``Instance normalization: The missing ingredient for fast stylization,'' \emph{arXiv preprint arXiv:1607.08022}, 2016.

\bibitem{wu2018group}
Y.~Wu and K.~He, ``Group normalization,'' in \emph{Proceedings of the European conference on computer vision (ECCV)}, 2018, pp. 3--19.

\bibitem{kalayeh2019training}
M.~M. Kalayeh and M.~Shah, ``Training faster by separating modes of variation in batch-normalized models,'' \emph{IEEE transactions on pattern analysis and machine intelligence}, vol.~42, no.~6, pp. 1483--1500, 2019.

\bibitem{Dempster77maximumlikelihood}
A.~P. Dempster, N.~M. Laird, and D.~B. Rubin, ``Maximum likelihood from incomplete data via the {EM} algorithm,'' \emph{JOURNAL OF THE ROYAL STATISTICAL SOCIETY, SERIES B}, vol.~39, no.~1, pp. 1--38, 1977.

\bibitem{cifar10_datasets}
A.~Krizhevsky, V.~Nair, and G.~Hinton, ``{CIFAR}-10 (canadian institute for advanced research),'' 2009.

\bibitem{cifar100_datasets}
------, ``{CIFAR}-100 (canadian institute for advanced research),'' 2009.

\bibitem{le2015tiny}
Y.~Le and X.~Yang, ``Tiny imagenet visual recognition challenge,'' \emph{CS 231N}, vol.~7, no.~7, p.~3, 2015.

\bibitem{mnist_datasets}
Y.~LeCun and C.~Cortes, ``{MNIST} handwritten digit database,'' 2010.

\bibitem{sermanet2012convolutional}
P.~Sermanet, S.~Chintala, and Y.~LeCun, ``Convolutional neural networks applied to house numbers digit classification,'' in \emph{Proceedings of the 21st international conference on pattern recognition (ICPR2012)}.\hskip 1em plus 0.5em minus 0.4em\relax IEEE, 2012, pp. 3288--3291.

\bibitem{lecun1998gradient}
Y.~LeCun, L.~Bottou, Y.~Bengio, and P.~Haffner, ``Gradient-based learning applied to document recognition,'' \emph{Proceedings of the IEEE}, vol.~86, no.~11, pp. 2278--2324, 1998.

\bibitem{relu}
V.~Nair and G.~E. Hinton, ``Rectified linear units improve restricted boltzmann machines,'' \emph{Proceedings of the 27th International Conference on Machine Learning (ICML-10)}, 2010.

\bibitem{bishop2006pattern}
C.~M. Bishop, \emph{Pattern Recognition and Machine Learning}.\hskip 1em plus 0.5em minus 0.4em\relax Springer, 2006.

\bibitem{loshchilov2017fixing}
I.~Loshchilov and F.~Hutter, ``Fixing weight decay regularization in adam,'' \emph{arXiv preprint arXiv:1711.05101}, 2017.

\bibitem{huang2017densely}
G.~Huang, Z.~Liu, L.~Van Der~Maaten, and K.~Q. Weinberger, ``Densely connected convolutional networks,'' in \emph{Proceedings of the IEEE conference on computer vision and pattern recognition}, 2017, pp. 4700--4708.

\bibitem{bengio2013advances}
Y.~Bengio, N.~Boulanger-Lewandowski, and R.~Pascanu, ``Advances in optimizing recurrent networks,'' in \emph{2013 IEEE international conference on acoustics, speech and signal processing}.\hskip 1em plus 0.5em minus 0.4em\relax IEEE, 2013, pp. 8624--8628.

\bibitem{farahani2021brief}
A.~Farahani, S.~Voghoei, K.~Rasheed, and H.~R. Arabnia, ``A brief review of domain adaptation,'' \emph{Advances in Data Science and Information Engineering: Proceedings from ICDATA 2020 and IKE 2020}, pp. 877--894, 2021.

\bibitem{berthelot2021adamatch}
D.~Berthelot, R.~Roelofs, K.~Sohn, N.~Carlini, and A.~Kurakin, ``Adamatch: A unified approach to semi-supervised learning and domain adaptation,'' \emph{arXiv preprint arXiv:2106.04732}, 2021.

\bibitem{keras_adamatch}
S.~Paul, ``Unifying semi-supervised learning and unsupervised domain adaptation with adamatch,'' 2019, https://github.com/keras-team/keras-io/tree/master.

\bibitem{zagoruyko2016wide}
S.~Zagoruyko and N.~Komodakis, ``Wide residual networks,'' \emph{arXiv preprint arXiv:1605.07146}, 2016.

\bibitem{kingma2014adam}
D.~P. Kingma and J.~Ba, ``Adam: A method for stochastic optimization,'' \emph{arXiv preprint arXiv:1412.6980}, 2014.

\end{thebibliography}
\end{document}